\documentclass{bmvc2k}

\title{Fine-Pruning: Joint Fine-Tuning and Compression of a Convolutional Network with Bayesian Optimization}

\addauthor{Frederick Tung}{ftung@sfu.ca}{1}
\addauthor{Srikanth Muralidharan}{smuralid@sfu.ca}{1}
\addauthor{Greg Mori}{mori@cs.sfu.ca}{1}

\addinstitution{
 School of Computing Science\\
 Simon Fraser University\\
 Burnaby, BC, Canada
}

\runninghead{Tung et al.}{Fine-Pruning: Joint Fine-Tuning and Compression}

\usepackage{amssymb}
\usepackage{algorithm}
\usepackage{algorithmic}

\begin{document}

\maketitle

\begin{abstract}
When approaching a novel visual recognition problem in a specialized image domain, a common strategy is to start with a pre-trained deep neural network and fine-tune it to the specialized domain. If the target domain covers a smaller visual space than the source domain used for pre-training (e.g. ImageNet), the fine-tuned network is likely to be over-parameterized. However, applying network pruning as a post-processing step to reduce the memory requirements has drawbacks: fine-tuning and pruning are performed independently; pruning parameters are set once and cannot adapt over time; and the highly parameterized nature of state-of-the-art pruning methods make it prohibitive to manually search the pruning parameter space for deep networks, leading to coarse approximations. We propose a principled method for jointly fine-tuning and compressing a pre-trained convolutional network that overcomes these limitations. Experiments on two specialized image domains (remote sensing images and describable textures) demonstrate the validity of the proposed approach.
\end{abstract}

\section{Introduction}
\label{sec:intro}

\begin{figure}[t]
\centering
\includegraphics[scale=0.45, trim=10 450 280 10, clip=true]{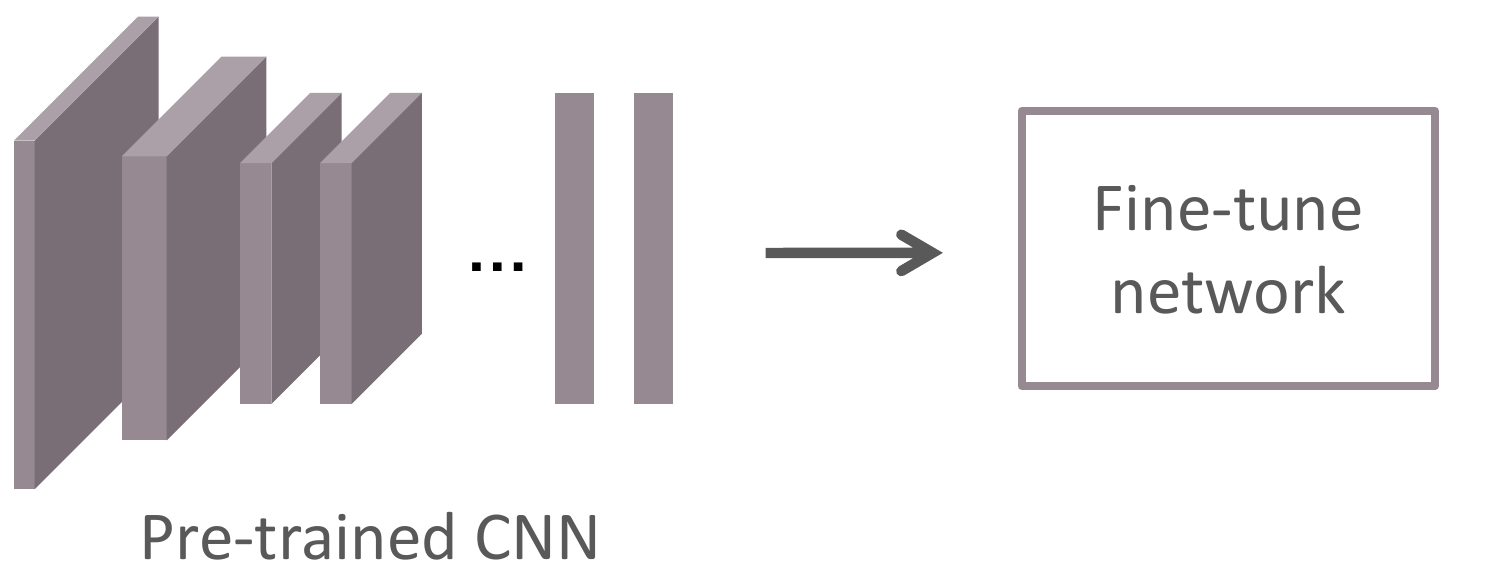} \\
(a) \\
\includegraphics[scale=0.45, trim=10 270 100 195, clip=true]{pull.pdf} \\
(b) \\
\includegraphics[scale=0.45, trim=10 20 60 390, clip=true]{pull.pdf} \\
(c)
\caption{
Consider the task of training a deep convolutional neural network on a specialized image domain (e.g. remote sensing images). (a) The conventional approach starts with a network pre-trained on a large, generic dataset (e.g. ImageNet) and fine-tunes it to the specialized domain. (b) Since the specialized domain spans a narrower visual space, the fine-tuned network is likely to be over-parameterized and can be compressed. A natural way to achieve this is to perform network pruning after fine-tuning, however this approach has limitations (see discussion in Section \ref{sec:intro}). (c) Fine-pruning addresses these limitations by jointly fine-tuning and compressing the pre-trained network in an iterative process. Each iteration consists of network fine-tuning, pruning module adaptation, and network pruning.}
\label{fig:pull}
\end{figure}

Convolutional neural networks (CNNs) have been widely adopted for visual analysis tasks such as image classification \cite{krizhevskyetal2012,simonyanzisserman2015}, object detection \cite{liuetal2016eccv,redmonetal2016}, action recognition \cite{simonyanzisserman2014,tranetal2015}, place recognition \cite{arandjelovicetal2016,zhouetal2014}, 3D shape classification \cite{johnetal2016,qietal2016}, image colorization \cite{zhangetal2016}, and camera pose estimation \cite{kendalletal2015}.
CNNs learn rich image and video representations that have been shown to generalize well across vision tasks.

When faced with a recognition task in a novel domain or application, a common strategy is to start with a CNN pre-trained on a large dataset, such as ImageNet \cite{ilsvrc}, and fine-tune the network to the new task (Fig.~\ref{fig:pull}a). 
Fine-tuning involves adapting the structure of the existing network to enable the new task, while re-using the pre-trained weights for the unmodified layers of the network. For example, a common and simple form of fine-tuning involves replacing the final fully-connected layer of the pre-trained CNN, which has an output dimensionality based on the pre-training dataset (e.g. 1000 dimensions for ImageNet), with a new fully-connected layer with a dimensionality that matches the target dataset. 

Fine-tuning allows powerful learned representations to be transferred to novel domains. Typically, we fine-tune complex network architectures that have been pre-trained on large databases containing millions of images. For example, we may fine-tune AlexNet \cite{krizhevskyetal2012} pre-trained on ImageNet's 1.2 million images (61 million parameters). In this way, we adapt these complex architectures to smaller and more specialized domains, such as remote sensing images. However, the specialized domain may not span the full space of natural images on which the original network was pre-trained. This suggests that the network architecture may be over-parameterized, and therefore inefficient in terms of memory and power consumption, with respect to the more constrained novel domain, in which a much more lightweight network would suffice for good performance. In applications with tight constraints on memory and power, such as mobile devices or robots, a more lightweight network with comparable classification accuracy may be valuable.

Given a fine-tuned network, a straightforward way to obtain a more lightweight network is to perform network pruning \cite{guoetal2016, hanetal2016, srinivasbabu2015} (Fig.~\ref{fig:pull}b). However, this strategy has drawbacks: (1) the fine-tuning and pruning operations are performed independently; (2)
the pruning parameters are set once and cannot adapt after training has started; and (3) since state-of-the-art pruning methods are highly parameterized, manually searching for good pruning hyperparameters is often prohibitive for deep networks, leading to coarse pruning strategies (e.g. pruning convolutional and fully connected layers separately \cite{guoetal2016}).

We propose a novel process called \emph{fine-pruning} (Fig.~\ref{fig:pull}c) that addresses these limitations:

\begin{enumerate}

\item Fine-pruning obtains a lightweight network specialized to a target domain by jointly fine-tuning and compressing the pre-trained network. The compatibility between the target domain and the pre-training domain is not normally known in advance (e.g. how similar are remote sensing images to ImageNet?), making it difficult to determine a priori how effective knowledge transfer will be, how aggressively compression can be applied, and where compression efforts should be focused. The knowledge transfer and network compression processes are linked and inform each other in fine-pruning.

\item Fine-pruning applies a principled adaptive network pruning strategy guided by Bayesian optimization, which automatically adapts the layer-wise pruning parameters over time as the network changes. For example, the Bayesian optimization controller might learn and execute a gradual pruning strategy in which network pruning is performed conservatively and fine-tuning restores the original accuracy in each iteration; or the controller might learn to prune aggressively at the outset and reduce the compression in later iterations (e.g. by splicing connections \cite{guoetal2016}) to recover accuracy.

\item Bayesian optimization enables efficient exploration of the pruning hyperparameter space, allowing all layers in the network to be considered together when making pruning decisions.

\end{enumerate}

\section{Related Work}
\label{sec:related}

\textbf{Network pruning.} Network pruning refers to the process of reducing the number of weights (connections) in a pre-trained neural network. The motivation behind this process is to make neural networks more compact and energy efficient for operation on resource constrained devices such as mobile phones. Network pruning can also improve network generalization by reducing overfitting. The earliest methods \cite{hassibistork1992,lecunetal1990} prune weights based on the second-order derivatives of the network loss. Data-free parameter pruning \cite{srinivasbabu2015} provides a data-independent method for discovering and removing entire neurons from the network. Deep compression \cite{hanetal2016} integrates the complementary techniques of weight pruning, scalar quantization to encode the remaining weights with fewer bits, and Huffman coding. Dynamic network surgergy \cite{guoetal2016} iteratively prunes and splices network weights. The novel splicing operation allows previously pruned weights to be reintroduced. Weights are pruned or spliced based on thresholding their absolute value. All weights, including pruned ones, are updated during backpropagation.

\noindent \textbf{Other network compression strategies.} Network pruning is one way to approach neural network compression. Other effective strategies include weight binarization \cite{courbariauxetal2015,rastegarietal2016}, architectural improvements \cite{iandolaetal2016}, weight quantization \cite{hanetal2016}, sparsity constraints \cite{lebedevlempitsky2016,zhouetal2016}, guided knowledge distillation \cite{hintonetal2015,romeroetal2015}, and replacement of fully connected layers with structured projections \cite{chengetal2015,moczulskietal2016,yangetal2015}.
Many of these network compression methods can train compact neural networks from scratch, or compress pre-trained networks for testing in the same domain.
However, since they assume particular types of weights, mimic networks trained in the same domain, or modify the network structure, most of these methods are not easily extended to the task of fine-tuning a pre-trained network to a specialized domain.

In this paper, we consider joint fine-tuning and network pruning in the context of transferring the knowledge of a pre-trained network to a smaller and more specialized visual recognition task. 
Previous approaches for compressing pre-trained neural networks aim to produce a compact network that performs as well as the original network \emph{on the dataset on which the network was originally trained}. In contrast, our focus is on the fine-tuning or \emph{transfer learning} problem of producing a compact network for a small, specialized target dataset, given a network pre-trained on a large, generic dataset such as ImageNet.
Our approach does not require the source dataset (e.g. ImageNet) on which the original network was trained.

\section{Method}
\label{sec:method}

Each fine-pruning iteration comprises three steps: fine-tuning, adaptation of the pruning module, and network pruning (Fig.~\ref{fig:pull}c). Fine-pruning can accommodate any parameterized network pruning module. In our experiments, we use the state-of-the-art dynamic network surgery method \cite{guoetal2016} for the network pruning module, but fine-pruning does not assume a particular pruning method.
Pruning module adaptation is guided by a Bayesian optimization \cite{gardneretal2014, snoeketal2012} controller, which enables an efficient search of the joint pruning parameter space, learning from the outcomes of previous exploration. This controller allows the pruning behaviour to change over time as connections are removed or formed.

Bayesian optimization is a general framework for solving global minimization problems involving blackbox objective functions:
\begin{equation}
\min_\mathbf{x} \; \ell(\mathbf{x}) \,, 
\label{eq:bo}
\end{equation}

\noindent where $\ell$ is a blackbox objective function that is typically expensive to evaluate, non-convex, may not be expressed in closed form, and may not be easily differentiable \cite{wangetal2013}. Eq.~\ref{eq:bo} is minimized by constructing a probabilistic model for $\ell$ to determine the most promising candidate $\mathbf{x}^*$ to evaluate next. Each iteration of Bayesian optimization involves selecting the most promising candidate $\mathbf{x}^*$, evaluating $\ell(\mathbf{x}^*)$, and using the data pair $(\mathbf{x}^*, \ell(\mathbf{x}^*))$ to update the probabilistic model for $\ell$.

In our case, $\mathbf{x}$ is a set of pruning parameters. For example, if the network pruning module is deep compression \cite{hanetal2016}, $\mathbf{x}$ consists of the magnitude thresholds used to remove weights; if the network pruning module is dynamic network surgery \cite{guoetal2016}, $\mathbf{x}$ consists of magnitude thresholds as well as cooling function hyperparameters that control how often the pruning mask is updated. We define $\ell$ by 
\begin{equation}
\ell(\mathbf{x}) = \varepsilon(\mathbf{x}) - \lambda \cdot s(\mathbf{x}) \,,
\label{eq:l}
\end{equation}

\noindent where $\varepsilon(\mathbf{x})$ is the top-1 error on the held-out validation set obtained by pruning the network according to the parameters $\mathbf{x}$ and then fine-tuning; $s(\mathbf{x})$ is the sparsity (proportion of pruned connections) of the pruned network obtained using the parameters $\mathbf{x}$; and $\lambda$ is an importance weight that balances accuracy and sparsity, which is set by held-out validation (we set $\lambda$ to maximize the achieved compression rate while maintaining the held-out validation error within a tolerance percentage, e.g. 2\%).

\begin{algorithm}[t]
\caption{Fine-Pruning}
\label{alg:summary}
\begin{algorithmic}[1]
\REQUIRE Pre-trained convolutional network, importance weight $\lambda$
\STATE Fine-tune network \COMMENT{$\triangleright$ Fig. \ref{fig:pull}a}
\REPEAT
\REPEAT[$\triangleright$ Bayesian optimization controller] 
\STATE Select next candidate parameters to evaluate as $\mathbf{x}^* = \operatorname*{arg\,max}_{\hat{\mathbf{x}}} \text{EI}(\hat{\mathbf{x}}) $
\STATE Evaluate $\ell(\mathbf{x}^*)$ 
\STATE Update Gaussian process model using $(\mathbf{x}^*, \ell(\mathbf{x}^*))$
\UNTIL{converged or maximum iterations of Bayesian optimization reached}
\STATE Prune network using best $\mathbf{x}^*$ found
\STATE Fine-tune network
\UNTIL{converged or maximum iterations of fine-pruning reached}
\end{algorithmic}
\end{algorithm}

We model the objective function as a Gaussian process \cite{rasmussenwilliams2006}. A Gaussian process is an uncountable set of random variables, any finite subset of which is jointly Gaussian. Let $\ell \sim \mathcal{GP}(\mu(\cdot), k(\cdot,\cdot))$, where $\mu(\cdot)$ is a mean function and $k(\cdot,\cdot)$ is a covariance kernel such that
\begin{align}
\mu(\mathbf{x}) &= \mathbb{E}\left[\ell(\mathbf{x})\right] \,, \\ \nonumber
k(\mathbf{x},\mathbf{x'}) &= \mathbb{E}\left[(\ell(\mathbf{x})-\mu(\mathbf{x}))(\ell(\mathbf{x'}) - \mu(\mathbf{x'}))\right] \,.
\end{align}

\noindent Given inputs $\mathbf{X} = \{\mathbf{x}_1, \mathbf{x}_2, ..., \mathbf{x}_n\}$ and function evaluations $\ell(\mathbf{X}) = \{\ell(\mathbf{x}_1), \ell(\mathbf{x}_2), ..., \ell(\mathbf{x}_n)\}$, the posterior belief of $\ell$ at a novel candidate $\hat{\mathbf{x}}$ can be computed in closed form. In particular,

\begin{equation}
\tilde{\ell}(\hat{\mathbf{x}}) \sim \mathcal{N}\left( \tilde{\mu}_\ell(\hat{\mathbf{x}}), \tilde{\Sigma}^2_\ell(\hat{\mathbf{x}})\right) \,,
\end{equation}

\noindent where 
\begin{align}
\tilde{\mu}_\ell(\hat{\mathbf{x}}) &= \mu(\hat{\mathbf{x}}) + k(\hat{\mathbf{x}}, \mathbf{X})k(\mathbf{X}, \mathbf{X})^{-1}(\ell(\mathbf{X}) - \mu(\mathbf{X})) \,, \\ \nonumber
\tilde{\Sigma}^2_\ell(\hat{\mathbf{x}}) &= k(\hat{\mathbf{x}}, \hat{\mathbf{x}}) - k(\hat{\mathbf{x}}, \mathbf{X})k(\mathbf{X}, \mathbf{X})^{-1}k(\mathbf{X}, \hat{\mathbf{x}}) \,.
\end{align}

\noindent The implication of the closed form solution is that, given a collection of parameters and the objective function evaluated at those parameters, we can efficiently predict the posterior at unevaluated parameters.

To select the most promising candidate to evaluate next, we use the expected improvement criterion. Let $\mathbf{x}^+$ denote the best candidate evaluated so far. The expected improvement of a candidate $\hat{\mathbf{x}}$ is defined as

\begin{equation}
\text{EI}(\hat{\mathbf{x}}) = \mathbb{E}\left[\max\left\{0, \ell(\mathbf{x}^+) - \tilde{\ell}(\hat{\mathbf{x}})\right\}\right] ,
\end{equation}

\noindent For a Gaussian process, the expected improvement of a candidate can also be efficiently computed in closed form. Specifically,
\begin{align}
\text{EI}(\hat{\mathbf{x}}) &= \tilde{\Sigma}_\ell(\hat{\mathbf{x}})(Z\Phi(Z) + \phi(Z)) \,, \\ \nonumber
Z &= \frac{\tilde{\mu}_\ell(\hat{\mathbf{x}})-\ell(\mathbf{x}^+)}{\tilde{\Sigma}_\ell(\hat{\mathbf{x}})} \,,
\end{align}

\noindent where $\Phi$ is the standard normal cumulative distribution function and $\phi$ is the standard normal probability density function. For a more detailed discussion on Gaussian processes and Bayesian optimization, we refer the interested reader to \cite{gardneretal2014}, \cite{rasmussenwilliams2006}, and \cite{snoeketal2012}. We use the publicly available code of \cite{gardneretal2014} and \cite{rasmussenwilliams2006} in our implementation.

The complete fine-pruning process is summarized in Algorithm \ref{alg:summary}.

\section{Experiments}

\begin{figure}
\centering
\includegraphics[scale=0.2]{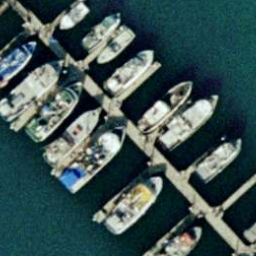} 
\includegraphics[scale=0.2]{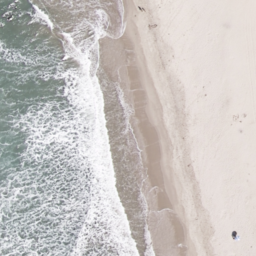}
\includegraphics[scale=0.2]{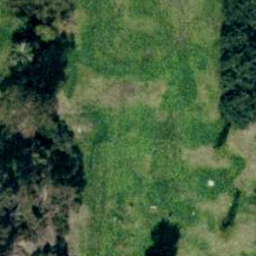}
\includegraphics[scale=0.2]{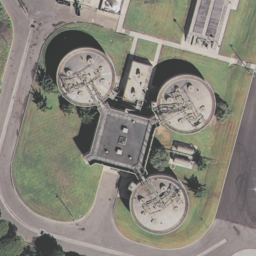}
\includegraphics[scale=0.2]{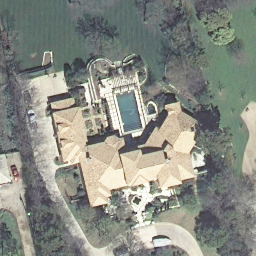}
\includegraphics[scale=0.2]{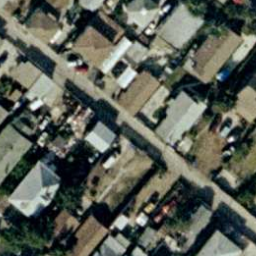}\\
(a) \\
\includegraphics[height=18mm]{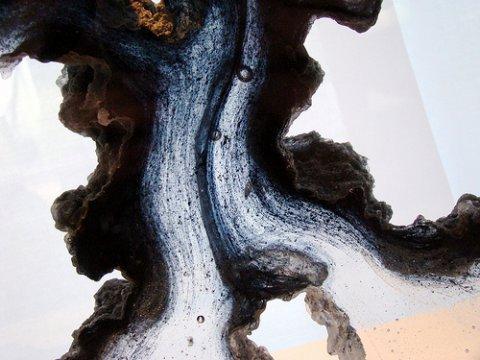} 
\includegraphics[height=18mm]{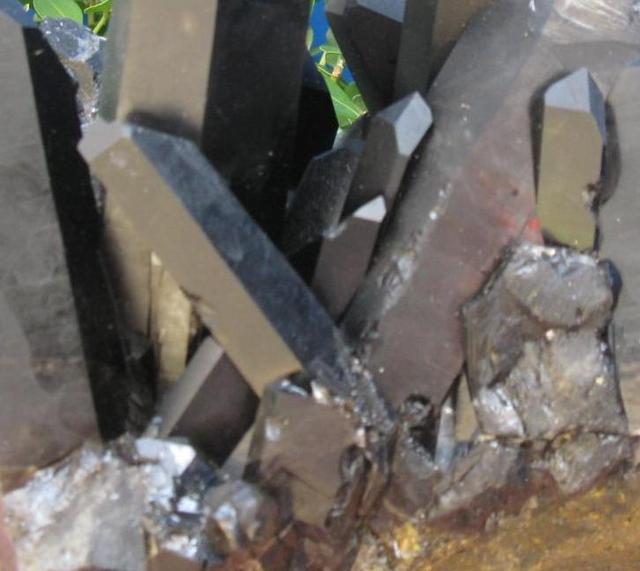}
\includegraphics[height=18mm]{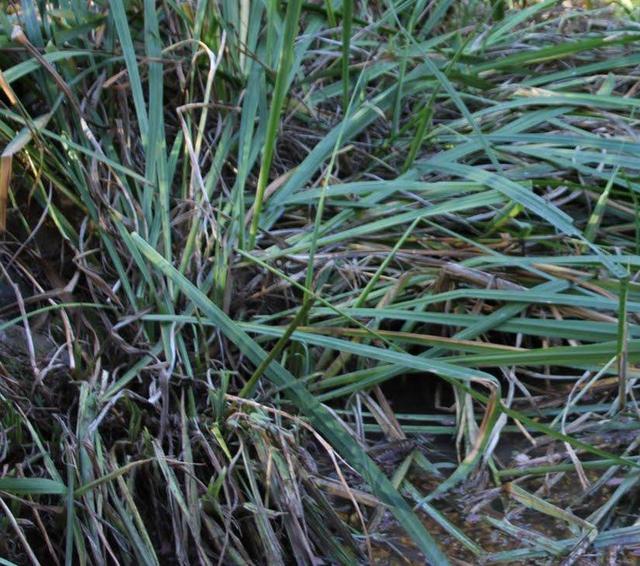}
\includegraphics[height=18mm]{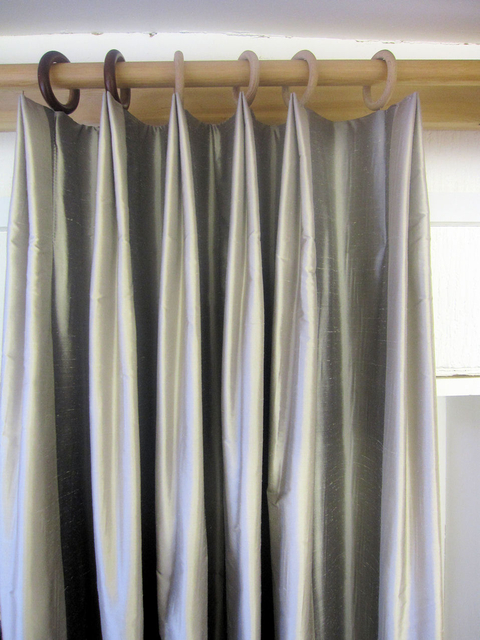}
\includegraphics[height=18mm]{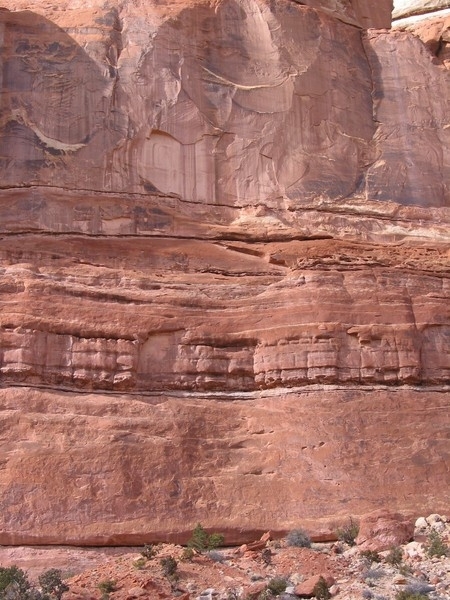}
\includegraphics[height=18mm]{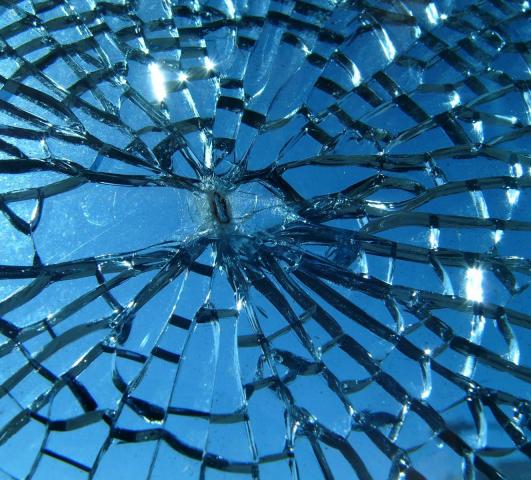}
\\
(b)
\caption{Sample images from the two specialized domain datasets used in our experiments: (a) Remote sensing images from the UCMerced Land Use Dataset \cite{yangetal2010}; (b) Texture images from the Describable Textures Dataset \cite{cimpoietal2014}.}
\label{fig:datasets}
\end{figure}

\textbf{Datasets.} We performed experiments on two specialized image domains: 
\begin{itemize}
\item Remote sensing images: The UCMerced Land Use Dataset \cite{yangetal2010} is composed of public domain aerial orthoimagery from the United States Geological Survey. The dataset covers 21 land-use classes, such as agricultural, dense residential, golf course, and harbor. Each land-use class is represented by 100 images. We randomly split the images into 50\% for training, 25\% for held-out validation, and 25\% for testing.
\item Describable textures: The Describable Textures Dataset \cite{cimpoietal2014} was introduced as part of a study in estimating human-describable texture attributes from images, which can then be used to improve tasks such as material recognition and description. The dataset consists of 5,640 images covering 47 human-describable texture attributes, such as blotchy, cracked, crystalline, fibrous, and pleated. We use the ten provided training, held-out validation, and testing splits.
\end{itemize}
Fig.~\ref{fig:datasets} shows examples of images from the two datasets.

\noindent \textbf{Baselines.} We compare fine-pruning with a fine-tuning only baseline (Fig. \ref{fig:pull}a) as well as independent fine-tuning followed by pruning (Fig. \ref{fig:pull}b), which for brevity we will refer to as the independent baseline. All experiments start from an ImageNet-pretrained AlexNet \cite{krizhevskyetal2012}. For a controlled comparison, we run the same state-of-the-art pruning method, dynamic network surgery \cite{guoetal2016}, in both the independent baseline and fine-pruning. In the original dynamic network surgery paper \cite{guoetal2016}, the authors prune convolutional and fully connected layers separately due to the prohibitive complexity of manually searching for layer-wise pruning parameters. To more fairly illustrate the benefit of fine-pruning, we set layer-wise pruning parameters for dynamic network surgery in the independent baseline using Bayesian optimization as well.

\begin{table}
\centering
\begin{tabular}{p{4cm}cccc}
\hline
\noalign{\smallskip}
& Accuracy & Accuracy & Parameters & Compression \\
& (Val.) & (Test) & & Rate \\
\noalign{\smallskip}
\hline
\noalign{\smallskip}\noalign{\smallskip}
\multicolumn{4}{l}{UCMerced Land Use Dataset \cite{yangetal2010}} \\ 
\noalign{\smallskip}
\hline
\noalign{\smallskip}
Fine-tuning only (Fig. \ref{fig:pull}a) & 94.7\% & 94.3\% & 57.0 M & -- \\
Independent fine-tuning and pruning (Fig. \ref{fig:pull}b) & 92.7$\pm$0.7\% & 93.8$\pm$0.7\% & 1.78$\pm$0.41 M & 31.9 $\times$ \\
Fine-pruning (Fig. \ref{fig:pull}c) & 92.5$\pm$0.9\% & 94.1$\pm$0.6\% & 1.17$\pm$0.39 M & \textbf{48.8 $\times$} \\
\noalign{\smallskip}
\hline
\noalign{\smallskip}\noalign{\smallskip}
\multicolumn{4}{l}{Describable Textures Dataset \cite{cimpoietal2014}} \\ 
\noalign{\smallskip}
\hline
\noalign{\smallskip}
Fine-tuning only (Fig. \ref{fig:pull}a) & 53.5$\pm$0.8\% & 53.7$\pm$0.9\% & 57.1 M & -- \\
Independent fine-tuning and pruning (Fig. \ref{fig:pull}b) & 52.8$\pm$1.2\% & 53.4$\pm$1.5\% & 3.62$\pm$0.54 M & 15.8 $\times$ \\
Fine-pruning (Fig. \ref{fig:pull}c) & 53.0$\pm$0.9\% & 52.8$\pm$0.8\% & 2.41$\pm$0.68 M & \textbf{23.7 $\times$} \\
\noalign{\smallskip}
\hline
\noalign{\medskip}
\end{tabular}
\caption{Experimental results on two specialized image domains: remote sensing images and describable textures. All experiments start with ImageNet-pretrained AlexNet \cite{krizhevskyetal2012} and use the state-of-the-art dynamic network surgery method \cite{guoetal2016} for network pruning. For a fair comparison, the pruning parameters in the independent fine-tuning and pruning baseline are also tuned by Bayesian optimization. We average results over ten runs on the remote sensing dataset and the ten provided splits on the describable textures dataset.} 
\label{tab:comparison}
\end{table}

\noindent \textbf{Implementation details.} We set all parameters by held-out validation on the two datasets. The importance weight $\lambda$ is set to 1 on both datasets. We warm-start both the independent baseline and fine-pruning with identical parameters obtained by random search. Fine-pruning is run to convergence or to a maximum of 10 iterations. In each fine-pruning iteration, Bayesian optimization considers up to 50 candidates and network fine-tuning is performed with a fixed learning rate of 0.001 (the same learning policy used to obtain the initial fine-tuned network) to 10 epochs.

\noindent \textbf{Results.} Table \ref{tab:comparison} summarizes our experimental comparison of fine-tuning, independent fine-tuning and pruning, and fine-pruning, on the UCMerced Land Use and Describable Textures datasets.
On UCMerced Land Use, the independent baseline produces sparse networks with 1.78 million parameters on average over ten runs, representing a reduction in the number of weights by 31.9-fold, while maintaining the test accuracy within 1\% of the dense fine-tuned network. Fine-pruning achieves further improvements in memory efficiency, producing sparse networks of 1.17 million parameters on average, or a 48.8-fold reduction in the number of weights, while maintaining the test accuracy within 1\% of the dense fine-tuned network.
On Describable Textures, we average the results over the ten provided splits. Similar improvements are obtained on this harder dataset. The independent baseline reduces the number of weights by 15.8-fold while maintaining the test accuracy within 1\% of the dense fine-tuned network. Fine-pruning lifts the compression rate to 23.7-fold while maintaining the test accuracy within 1\% of the dense fine-tuned network.

\begin{figure}
\centering
\includegraphics[width=0.45\linewidth,clip=true,trim=180 280 190 280]{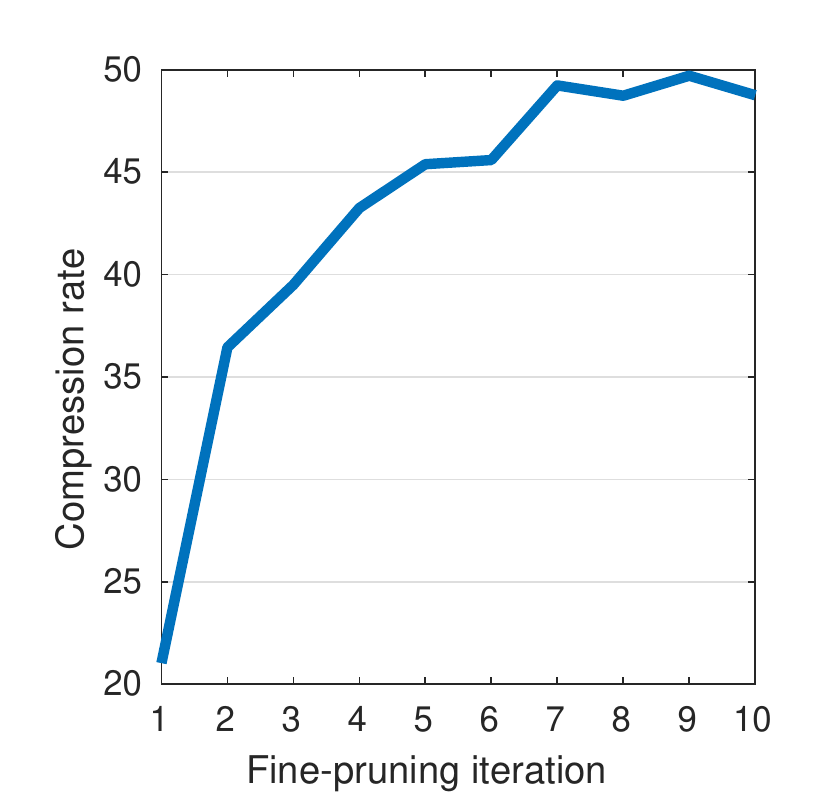}
\includegraphics[width=0.45\linewidth,clip=true,trim=180 280 190 280]{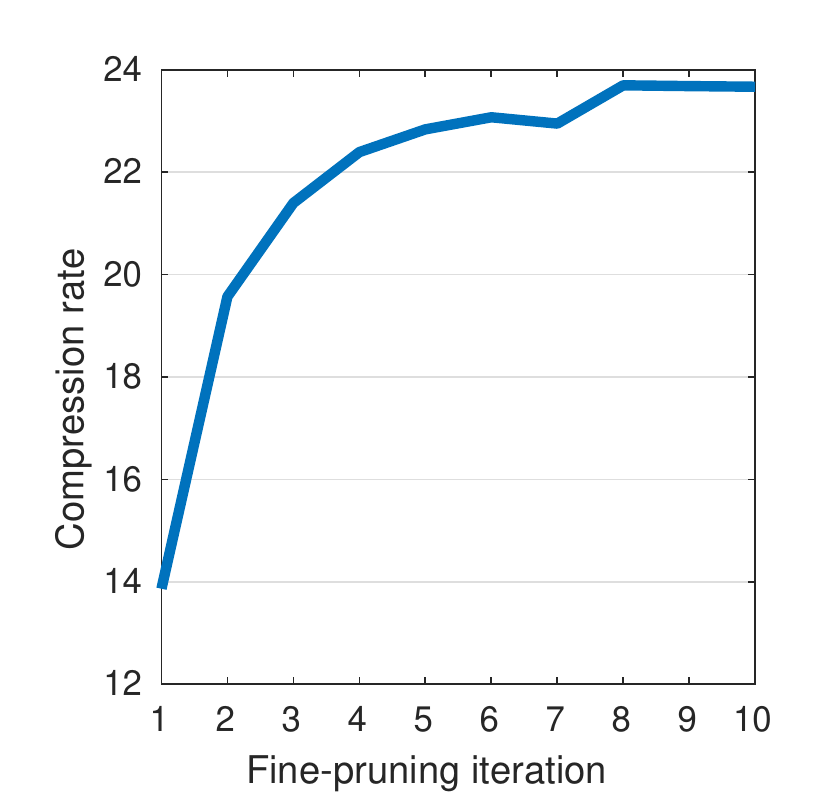} \\
(a) \hspace{5.5cm} (b)
\caption{Compression as a function of fine-pruning iteration. On both the (a) UCMerced Land Use Dataset and (b) Describable Textures Dataset, the pruning module adaptation, guided by Bayesian optimization, learns a policy of starting with a strong initial prune and tapering off in later iterations.}
\label{fig:compression_vs_iter}
\end{figure}

Fig. \ref{fig:compression_vs_iter} shows how the compression rate varies with the fine-pruning iteration. We observe that, on both datasets, the pruning module adaptation learns to start with a strong initial prune and then gradually increase pruning aggressiveness in later iterations until the network converges. 
This behavior can also be observed by examining the pruning parameters $\mathbf{x}^*$ selected by Bayesian optimization.

Table~\ref{tab:layerwise} illustrates the average number of weights layer by layer after fine-pruning for both datasets. We observe that the original fine-tuned networks in both cases are highly over-parameterized, and a significant reduction in memory can be obtained by fine-pruning. A large proportion of the original network parameters reside in the fully connected layers fc6 and fc7. Provided that the underlying network pruning module allows for pruning parameters to be set on an individual layer basis, our Bayesian optimization controller automatically learns to prioritize the compression of these layers because they have the largest influence on $s(\mathbf{x})$ in the objective function (Eq.~\ref{eq:l}). 

\begin{table}
\centering
\begin{tabular}{p{2cm}ccc}
\hline
\noalign{\smallskip}
 & Parameters: & Parameters: & Percentage \\
& Before & After & Pruned\\
\noalign{\smallskip}
\hline
\noalign{\smallskip}\noalign{\smallskip}
\multicolumn{4}{l}{UCMerced Land Use Dataset \cite{yangetal2010}} \\ 
\noalign{\smallskip}
\hline
\noalign{\smallskip}
conv1 & 35 K & 26 K & 26.1\%\\
conv2 & 307 K & 92 K & 70.2\%\\
conv3 & 885 K & 261 K & 70.5\%\\
conv4 & 664 K & 218 K & 67.2\%\\
conv5 & 443 K & 181 K & 59.1\%\\
fc6 & 37.8 M & 313 K & 99.2\%\\
fc7 & 16.8 M & 60 K & 99.6\%\\
fc8 & 86 K & 17 K & 80.3\%\\
\hline
total & 57.0 M & 1.17 M & 98.0\%\\
\noalign{\smallskip}
\hline
\noalign{\smallskip}\noalign{\smallskip}
\multicolumn{4}{l}{Describable Textures Dataset \cite{cimpoietal2014}} \\ 
\noalign{\smallskip}
\hline
\noalign{\smallskip}
conv1 & 35 K & 32 K & 8.3\%\\
conv2 & 307 K & 245 K & 20.4\%\\
conv3 & 885 K & 343 K & 61.3\% \\
conv4 & 664 K & 442 K & 33.4\%\\
conv5 & 443 K & 216 K & 51.2\% \\
fc6 & 37.8 M & 401 K & 98.9\% \\
fc7 & 16.8 M & 661 K & 96.1\% \\
fc8 & 193 K & 72 K & 62.4\% \\ \hline
total & 57.1 M & 2.41 M & 95.8\% \\
\noalign{\smallskip}
\hline
\noalign{\medskip}
\end{tabular}
\caption{Layer-wise compression results. Our Bayesian optimization controller automatically learns to prioritize the compression of the fc6 and fc7 layers, which have the most parameters.}
\label{tab:layerwise}
\end{table}

\section{Conclusion}

In this paper we have presented a joint process for network fine-tuning and compression that produces a memory-efficient network tailored to a specialized image domain. Our process is guided by a Bayesian optimization controller that allows pruning parameters to adapt over time to the characteristics of the changing network. Fine-pruning is general and can accommodate any parameterized network pruning algorithm. In future we plan to study whether our technique can be applied to provide better time efficiency as well. For example, structured sparsity may result in more significant time savings on a GPU than unstructured sparsity \cite{wenetal2016}. Specialized hardware engines \cite{hanetal2016eie} can also accelerate networks with unstructured sparsity while reducing energy consumption.

\noindent \textbf{Acknowledgements.} This work was supported by the Natural Sciences and Engineering Research Council of Canada.

\bibliography{fp}

\begin{thebibliography}{36}
\providecommand{\natexlab}[1]{#1}
\providecommand{\url}[1]{\texttt{#1}}
\expandafter\ifx\csname urlstyle\endcsname\relax
  \providecommand{\doi}[1]{doi: #1}\else
  \providecommand{\doi}{doi: \begingroup \urlstyle{rm}\Url}\fi

\bibitem[Arandjelovic et~al.(2016)Arandjelovic, Gronat, Torii, Pajdla, and
  Sivic]{arandjelovicetal2016}
R.~Arandjelovic, P.~Gronat, A.~Torii, T.~Pajdla, and J.~Sivic.
\newblock {NetVLAD}: {CNN} architecture for weakly supervised place
  recognition.
\newblock In \emph{IEEE Conference on Computer Vision and Pattern Recognition},
  2016.

\bibitem[Cheng et~al.(2015)Cheng, Yu, Feris, Kumar, Choudhary, and
  Chang]{chengetal2015}
Y.~Cheng, F.~X. Yu, R.~S. Feris, S.~Kumar, A.~Choudhary, and S.-F. Chang.
\newblock An exploration of parameter redundancy in deep networks with
  circulant projections.
\newblock In \emph{IEEE International Conference on Computer Vision}, 2015.

\bibitem[Cimpoi et~al.(2014)Cimpoi, Maji, Kokkinos, Mohamed, and
  Vedaldi]{cimpoietal2014}
M.~Cimpoi, S.~Maji, I.~Kokkinos, S.~Mohamed, and A.~Vedaldi.
\newblock Describing textures in the wild.
\newblock In \emph{IEEE Conference on Computer Vision and Pattern Recognition},
  2014.

\bibitem[Courbariaux et~al.(2015)Courbariaux, Bengio, and
  David]{courbariauxetal2015}
M.~Courbariaux, Y.~Bengio, and J.-P. David.
\newblock {BinaryConnect}: Training deep neural networks with binary weights
  during propagations.
\newblock In \emph{Advances in Neural Information Processing Systems}, 2015.

\bibitem[Gardner et~al.(2014)Gardner, Kusner, Xu, Weinberger, and
  Cunningham]{gardneretal2014}
J.~R. Gardner, M.~J. Kusner, Z.~Xu, K.~Q. Weinberger, and J.~P. Cunningham.
\newblock Bayesian optimization with inequality constraints.
\newblock In \emph{International Conference on Machine Learning}, 2014.

\bibitem[Guo et~al.(2016)Guo, Yao, and Chen]{guoetal2016}
Y.~Guo, A.~Yao, and Y.~Chen.
\newblock Dynamic network surgery for efficient {DNNs}.
\newblock In \emph{Advances in Neural Information Processing Systems}, 2016.

\bibitem[Han et~al.(2016{\natexlab{a}})Han, Liu, Mao, Pu, Pedram, Horowitz, and
  Dally]{hanetal2016eie}
S.~Han, X.~Liu, H.~Mao, J.~Pu, A.~Pedram, M.~A. Horowitz, and W.~J. Dally.
\newblock {EIE}: Efficient inference engine on compressed deep neural network.
\newblock In \emph{ACM/IEEE International Symposium on Computer Architecture},
  2016{\natexlab{a}}.

\bibitem[Han et~al.(2016{\natexlab{b}})Han, Mao, and Dally]{hanetal2016}
S.~Han, H.~Mao, and W.~J. Dally.
\newblock {Deep Compression}: Compressing deep neural networks with pruning,
  trained quantization and {Huffman} coding.
\newblock In \emph{International Conference on Learning Representations},
  2016{\natexlab{b}}.

\bibitem[Hassibi and Stork(1992)]{hassibistork1992}
B.~Hassibi and D.~G. Stork.
\newblock Second order derivatives for network pruning: optimal brain surgeon.
\newblock In \emph{Advances in Neural Information Processing Systems}, 1992.

\bibitem[Hinton et~al.(2015)Hinton, Vinyals, and Dean]{hintonetal2015}
G.~Hinton, O.~Vinyals, and J.~Dean.
\newblock Distilling the knowledge in a neural network.
\newblock arXiv:1503.02531, 2015.

\bibitem[Iandola et~al.(2016)Iandola, Han, Moskewicz, Ashraf, Dally, and
  Keutzer]{iandolaetal2016}
F.~N. Iandola, S.~Han, M.~W. Moskewicz, K.~Ashraf, W.~J. Dally, and K.~Keutzer.
\newblock {SqueezeNet}: {AlexNet}-level accuracy with 50x fewer parameters and
  {<0.5MB} model size.
\newblock arXiv:1602.07360, 2016.

\bibitem[Johns et~al.(2016)Johns, Leutenegger, and Davison]{johnetal2016}
E.~Johns, S.~Leutenegger, and A.~J. Davison.
\newblock Pairwise decomposition of image sequences for active multi-view
  recognition.
\newblock In \emph{IEEE Conference on Computer Vision and Pattern Recognition},
  2016.

\bibitem[Kendall et~al.(2015)Kendall, Grimes, and Cipolla]{kendalletal2015}
A.~Kendall, M.~Grimes, and R.~Cipolla.
\newblock {PoseNet}: A convolutional network for real-time 6-{DOF} camera
  relocalization.
\newblock In \emph{IEEE International Conference on Computer Vision}, 2015.

\bibitem[Krizhevsky et~al.(2012)Krizhevsky, Sutskever, and
  Hinton]{krizhevskyetal2012}
A.~Krizhevsky, I.~Sutskever, and G.~E. Hinton.
\newblock {ImageNet} classification with deep convolutional neural networks.
\newblock In \emph{Advances in Neural Information Processing Systems}, 2012.

\bibitem[Lebedev and Lempitsky(2016)]{lebedevlempitsky2016}
V.~Lebedev and V.~Lempitsky.
\newblock Fast {ConvNets} using group-wise brain damage.
\newblock In \emph{IEEE Conference on Computer Vision and Pattern Recognition},
  2016.

\bibitem[LeCun et~al.(1990)LeCun, Denker, and Solla]{lecunetal1990}
Y.~LeCun, J.~S. Denker, and S.~A. Solla.
\newblock Optimal brain damage.
\newblock In \emph{Advances in Neural Information Processing Systems}, 1990.

\bibitem[Liu et~al.(2016)Liu, Anguelov, Erhan, Szegedy, Reed, Fu, and
  Berg]{liuetal2016eccv}
W.~Liu, D.~Anguelov, D.~Erhan, C.~Szegedy, S.~Reed, C.-Y. Fu, and A.~C. Berg.
\newblock {SSD}: Single shot multibox detector.
\newblock In \emph{European Conference on Computer Vision}, 2016.

\bibitem[Moczulski et~al.(2016)Moczulski, Denil, Appleyard, and
  de~Freitas]{moczulskietal2016}
M.~Moczulski, M.~Denil, J.~Appleyard, and N.~de~Freitas.
\newblock {ACDC}: A structured efficient linear layer.
\newblock In \emph{International Conference on Learning Representations}, 2016.

\bibitem[Qi et~al.(2016)Qi, Su, Nie{\ss}ner, Dai, Yan, and Guibas]{qietal2016}
C.~R. Qi, H.~Su, M.~Nie{\ss}ner, A.~Dai, M.~Yan, and L.~J. Guibas.
\newblock Volumetric and multi-view {CNNs} for object classification on {3D}
  data.
\newblock In \emph{IEEE Conference on Computer Vision and Pattern Recognition},
  2016.

\bibitem[Rasmussen and Williams(2006)]{rasmussenwilliams2006}
C.~E. Rasmussen and C.~K.~I. Williams.
\newblock \emph{Gaussian Processes for Machine Learning}.
\newblock MIT Press, 2006.

\bibitem[Rastegari et~al.(2016)Rastegari, Ordonez, Redmon, and
  Farhadi]{rastegarietal2016}
M.~Rastegari, V.~Ordonez, J.~Redmon, and A.~Farhadi.
\newblock {XNOR-Net}: {ImageNet} classification using binary convolutional
  neural networks.
\newblock In \emph{European Conference on Computer Vision}, 2016.

\bibitem[Redmon et~al.(2016)Redmon, Divvala, Girshick, and
  Farhadi]{redmonetal2016}
J.~Redmon, S.~Divvala, R.~Girshick, and A.~Farhadi.
\newblock You only look once: unified, real-time object detection.
\newblock In \emph{IEEE Conference on Computer Vision and Pattern Recognition},
  2016.

\bibitem[Romero et~al.(2015)Romero, Ballas, Kahou, Chassang, Gatta, and
  Bengio]{romeroetal2015}
A.~Romero, N.~Ballas, S.~E. Kahou, A.~Chassang, C.~Gatta, and Y.~Bengio.
\newblock {FitNets}: hints for thin deep nets.
\newblock In \emph{International Conference on Learning Representations}, 2015.

\bibitem[Russakovsky et~al.(2014)Russakovsky, Deng, Su, Krause, Satheesh, Ma,
  Huang, Karpathy, Khosla, Bernstein, Berg, and Fei-Fei]{ilsvrc}
O.~Russakovsky, J.~Deng, H.~Su, J.~Krause, S.~Satheesh, S.~Ma, Z.~Huang,
  A.~Karpathy, A.~Khosla, M.~Bernstein, A.~C. Berg, and L.~Fei-Fei.
\newblock {ImageNet Large Scale Visual Recognition Challenge}.
\newblock arXiv:1409.0575, 2014.

\bibitem[Simonyan and Zisserman(2014)]{simonyanzisserman2014}
K.~Simonyan and A.~Zisserman.
\newblock Two-stream convolutional networks for action recognition in videos.
\newblock In \emph{Advances in Neural Information Processing Systems}, 2014.

\bibitem[Simonyan and Zisserman(2015)]{simonyanzisserman2015}
K.~Simonyan and A.~Zisserman.
\newblock Very deep convolutional networks for large-scale image recognition.
\newblock In \emph{International Conference on Learning Representations}, 2015.

\bibitem[Snoek et~al.(2012)Snoek, Larochelle, and Adams]{snoeketal2012}
J.~Snoek, H.~Larochelle, and R.~P. Adams.
\newblock Practical {Bayesian} optimization of machine learning algorithms.
\newblock In \emph{Advances in Neural Information Processing Systems}, 2012.

\bibitem[Srinivas and Babu(2015)]{srinivasbabu2015}
S.~Srinivas and R.~V. Babu.
\newblock Data-free parameter pruning for deep neural networks.
\newblock In \emph{British Machine Vision Conference}, 2015.

\bibitem[Tran et~al.(2015)Tran, Bourdev, Fergus, Torresani, and
  Paluri]{tranetal2015}
D.~Tran, L.~Bourdev, R.~Fergus, L.~Torresani, and M.~Paluri.
\newblock Learning spatiotemporal features with {3D} convolutional networks.
\newblock In \emph{IEEE International Conference on Computer Vision}, 2015.

\bibitem[Wang et~al.(2013)Wang, Zoghi, Hutter, Matheson, and
  de~Freitas]{wangetal2013}
Z.~Wang, M.~Zoghi, F.~Hutter, D.~Matheson, and N.~de~Freitas.
\newblock Bayesian optimization in high dimensions via random embeddings.
\newblock In \emph{International Joint Conference on Artificial Intelligence},
  2013.

\bibitem[Wen et~al.(2016)Wen, Wu, Wang, Chen, and Li]{wenetal2016}
W.~Wen, C.~Wu, Y.~Wang, Y.~Chen, and H.~Li.
\newblock Learning structured sparsity in deep neural networks.
\newblock In \emph{Advances in Neural Information Processing Systems}, 2016.

\bibitem[Yang and Newsam(2010)]{yangetal2010}
Y.~Yang and S.~Newsam.
\newblock Bag-of-visual-words and spatial extensions for land-use
  classification.
\newblock In \emph{ACM SIGSPATIAL International Conference on Advances in
  Geographic Information Systems}, 2010.

\bibitem[Yang et~al.(2015)Yang, Moczulski, Denil, de~Freitas, Smola, Song, and
  Wang]{yangetal2015}
Z.~Yang, M.~Moczulski, M.~Denil, N.~de~Freitas, A.~Smola, L.~Song, and Z.~Wang.
\newblock Deep fried convnets.
\newblock In \emph{IEEE International Conference on Computer Vision}, 2015.

\bibitem[Zhang et~al.(2016)Zhang, Isola, and Efros]{zhangetal2016}
R.~Zhang, P.~Isola, and A.~A. Efros.
\newblock Colorful image colorization.
\newblock In \emph{European Conference on Computer Vision}, 2016.

\bibitem[Zhou et~al.(2014)Zhou, Lapedriza, Xiao, Torralba, and
  Oliva]{zhouetal2014}
B.~Zhou, A.~Lapedriza, J.~Xiao, A.~Torralba, and A.~Oliva.
\newblock Learning deep features for scene recognition using places database.
\newblock In \emph{Advances in Neural Information Processing Systems}, 2014.

\bibitem[Zhou et~al.(2016)Zhou, Alvarez, and Porikli]{zhouetal2016}
H.~Zhou, J.~M. Alvarez, and F.~Porikli.
\newblock Less is more: towards compact {CNNs}.
\newblock In \emph{European Conference on Computer Vision}, 2016.

\end{thebibliography}
\end{document}